\documentclass[runningheads]{llncs}
\usepackage{graphicx}
\usepackage{amsmath,amssymb} 
\usepackage{color}
\usepackage[width=122mm,left=12mm,paperwidth=146mm,height=193mm,top=12mm,paperheight=217mm]{geometry}

\usepackage{enumitem}
\usepackage{tabu}
\usepackage{multirow}
\usepackage{color}
\usepackage{xcolor}
\usepackage{colortbl}
\usepackage{subcaption}

\begin{document}
\pagestyle{headings}
\mainmatter

\title{Automatically selecting inference algorithms\\for discrete energy minimisation}

\titlerunning{Automatically selecting inference algorithms for discrete energy minimisation}

\authorrunning{P. Henderson, V. Ferrari}

\author{Paul Henderson \& Vittorio Ferrari}
\institute{
  School of Informatics, University of Edinburgh\\
  \email{ \{p.m.henderson,vittorio.ferrari\}@ed.ac.uk }
}

\newcommand{\vitto}[1]{\textcolor{red}{[VF: #1]}}
\newcommand{\pmh}[1]{\textcolor{blue}{[PH: #1]}}

\newcommand{\sect}[1]{sec.~\ref{sec:#1}}
\newcommand{\fig}[1]{fig.~\ref{fig:#1}}
\newcommand{\tab}[1]{table~\ref{tab:#1}}
\newcommand{\eq}[1]{eq.~(\ref{eq:#1})}

\newcommand{\eg}{\textit{e.g.\ }}
\newcommand{\ie}{\textit{i.e.\ }}

\newcommand{\defaultarraystretch}{1.25}
\renewcommand{\arraystretch}{\defaultarraystretch}

\definecolor{grey}{gray}{0.5}

\renewcommand{\paragraph}[1]{\vskip 2pt\par\noindent\textbf{#1}\hskip 0.5em} 

\newcommand{\afterfig}{}

\maketitle

\begin{abstract}

Minimisation of discrete energies defined over factors is an important problem in computer vision, and a vast number of MAP inference algorithms have been proposed.
Different inference algorithms perform better on factor graph models (GMs) from different underlying problem classes, and in general it is difficult to know which algorithm will yield the lowest energy for a given GM.
To mitigate this difficulty, survey papers~\cite{kolmogorov06eccv,szeliski:pami08,kappes15ijcv} advise the practitioner on what algorithms perform well on what classes of models.
We take the next step forward, and present a technique to automatically select the best inference algorithm for an input GM. We validate our method experimentally on an extended version of the OpenGM2 benchmark~\cite{kappes15ijcv},
containing a diverse set of vision problems. On average, our method selects an inference algorithm yielding labellings with 96\% of variables the same as the best available algorithm. 

\end{abstract}

\section{Introduction}\label{sec:introduction}

Minimisation of discrete energies defined over factors is an important problem in computer vision and other fields such as bioinformatics, with many algorithms proposed in the literature to solve such problems~\cite{kappes15ijcv}.
These models arise from many different underlying \textit{problem classes}; in vision, typical examples are stereo matching, semantic segmentation, and texture reconstruction, each of which yields models with very different characteristics,
making different choices of minimisation algorithm preferable.

We consider factor graph models (GMs) defined by sets $V$ and $F$ of variables and factors respectively.
Each variable takes values in some discrete label-space, and each factor is a real-valued function on some subset of $V$, its clique, yielding an additive contribution to a global energy.
In this paper, we focus on algorithms to find the labelling of variables that minimises this global energy, the so-called {\em MAP inference} problem.
Different problem classes give rise to problem instances with different characteristics, such as size of cliques and number of variables, affecting which inference algorithms are best suited to them.

\begin{figure}
  \centering
  \includegraphics[width=0.6\textwidth]{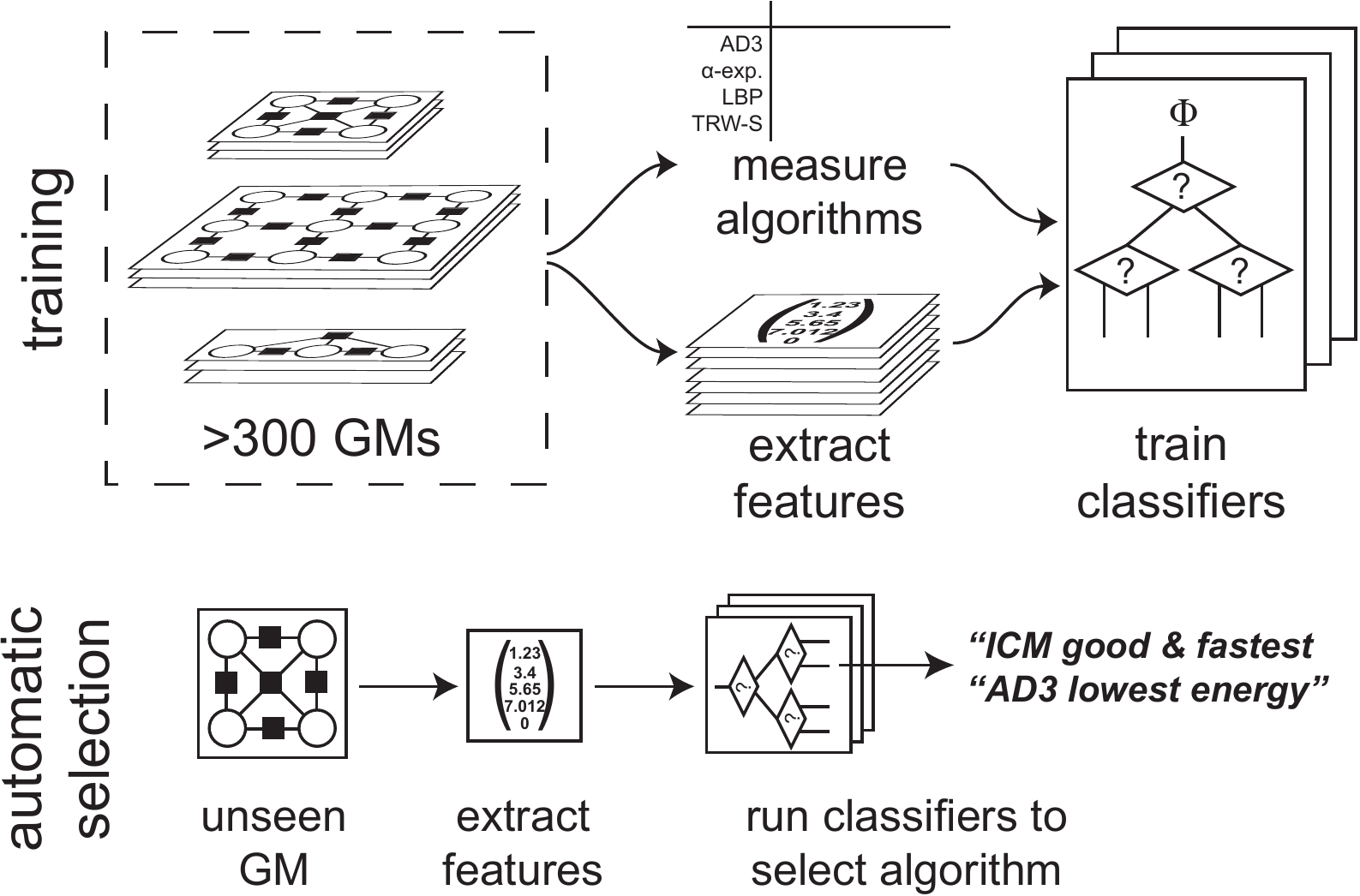}
  \caption{Our pipeline for automatic algorithm selection}
  \label{fig:overview}
  \afterfig
\end{figure}

The space of published inference algorithms is vast, with methods ranging from highly specialised to very general.
For example, message passing~\cite{Bishop-book} is widely applicable, but takes exponential time for large cliques, and may not converge.
Dual-space variants such as TRW-S~\cite{kolmogorov06pami} do guarantee convergence, but not necessarily to a global optimum.
$\alpha$-expansion~\cite{boykov01pami} and graph-cuts~\cite{greig89jrss} are better suited to models with dense connectivity, but require factors 
to take certain restricted forms,
while QPBO~\cite{rother:cvpr07} only works for binary models and may leave some variables unlabelled.
Algorithms solving the Wolfe dual~\cite{storvik00tip,guignard87mp,komodakis07iccv} such as \cite{kappes12cvpr,martins15jmlr} are applicable to models with arbitrary factors and labels,
but existing implementations for this generic setting tend to run more slowly.

Thus, when developing a new model, it may be difficult to decide what algorithm to use for inference.
Selecting a good algorithm for a given model requires extensive expertise about the landscape of existing algorithms and typically involves understanding the operational details of many of them.
Moreover, even for an expert who can choose which algorithm is best overall on a particular problem class, it may not be clear which is best for a particular \textit{instance}---certain problem classes are heterogeneous enough that different instances within them may be best solved by different algorithms (\sect{algorithms}). An alternative solution would be to run many algorithms on each input model and see which one performs best empirically. However, this would be computationally very expensive.

Recently studies appeared that evaluate a number of algorithms on various problems, comparing their performance~\cite{kappes15ijcv,szeliski:pami08,andres10dagm,kolmogorov06eccv,alahari10pami}.
These are intended to provide a `field guide' for the practitioner, suggesting which techniques are suited for which models.
In this paper, we take the next step forward and propose a technique to \textit{automatically} select which inference algorithm to run on an input problem instance (\sect{learnt-selection}).
We do so without requiring the user to have any knowledge of the applicability of different inference methods, and without the computational expense of running many algorithms.
Thus, our method is particularly suited for the vision practitioner with limited knowledge of inference, but who wishes to apply it to real-world problems.

Our method uses features extracted from the problem instance itself, to select inference algorithms according to two criteria relevant for the practitioner:
(1) the fastest algorithm reaching the lowest energy for that instance; or
(2) the fastest algorithm delivering a very similar labelling to the lowest energy one (\fig{overview}).
The features are designed to capture characteristics of the instance that affect algorithm applicability or performance, such as the clique sizes and connectivity structure (\sect{features}).
We train our selection models without human supervision, based on the results of running many algorithms over a large dataset of training problem instances.

We perform experiments (\sect{results}) on an extended version of the OpenGM2 benchmark~\cite{kappes15ijcv}, containing 344 problem instances drawn from 32 diverse classes (\sect{dataset}), and consider a pool of 15 inference algorithms drawn from the most prominent approaches (\sect{algorithms}).
The results show that on 69\% of problem instances our method selects the best algorithm. 
On average, the labels of 96\% of variables match that returned by the algorithm achieving the lowest energy.
Our automatic selector achieves these results over $88\times$ faster than the obvious alternative of running all algorithms and retaining the best solution.

\subsection{Related work}
\label{sec:related-work}

\paragraph{MAP inference.}
MAP inference algorithms can be split into several broad categories.
Graph-cuts~\cite{greig89jrss} is very efficient, but restricted to pairwise binary GMs with submodular factors.
It can be extended to more general models, such as by the move-making methods $\alpha$-expansion and $\alpha\beta$-swap~\cite{boykov01pami}, wherein a subset of variables change label at each iteration, or by transformations introducing auxiliary variables~\cite{ishikawa09cvpr,ishikawa11pami,fix11iccv}.
Alternatively, inference is naturally formulated as an integer linear program, which can be solved directly and optimally using off-the-shelf polyhedral solvers for small problems~\cite{kappes15ijcv}.
It can also be relaxed to a non-integer linear program (LP), which can be solved faster. However, it requires rounding the solution, which does not always yield the global optimum of the original problem.
Message-passing algorithms~\cite{Bishop-book,kschischang01tinft} have each variable/factor iteratively send to its neighbours messages encoding its current belief about each neighbour's min-marginals.
Tree-reweighted methods~\cite{kolmogorov06pami,wainwright:inftheo2005} use a message-passing formulation, but actually solve a Lagrangian dual of the LP, and can provide a certificate of optimality where relevant.
Other dual-decomposition methods~\cite{kappes12cvpr,martins15jmlr,sontag08uai} directly solve the Wolfe dual~\cite{guignard87mp,komodakis07iccv} to the LP, but by iteratively finding the MAP state of each clique (or other tractable subgraphs) instead of passing messages.
Our focus in this paper is not to introduce another inference algorithm, but to consider the meta-problem of learning to select what existing inference algorithm to apply to an input model; as such, we use many of the above algorithms in our framework (\sect{algorithms}).

\paragraph{Inferning.}
Our work is a form of {\em inferning}~\cite{inferning-icml13}, as it considers interactions between inference and learning.
A few such methods use learning to guide the inference process.
Unlike the hard-wired algorithms mentioned above, these approaches learn to adapt to the characteristics of a particular problem class.
Some operate by pruning the model during inference, by learning classifiers to remove labels from some variables~\cite{guillaumin13cvpr,conejo14nips}, or to remove certain factors from the model~\cite{stoyanov12icmlwks,roig13iccv}.
Others learn an optimal sequence of operations to perform during
message-passing inference~\cite{jiang13icmlwks}.
Our work operates at a higher level than these approaches.
Instead of incorporating learning into an algorithm to allow adaptation to a problem class, we instead learn to predict which of a fixed set of hard-wired algorithms is best to apply to a given problem instance.

\paragraph{Surveys on inference.}
The survey papers \cite{kolmogorov06eccv,szeliski:pami08,andres10dagm,alahari10pami,kappes15ijcv} evaluate a number of algorithms on various problems, comparing their performance. \cite{kolmogorov06eccv} focuses on  stereo matching  and considers highly-connected grid models defined on pixels with unary and pairwise factors only. It evaluates three inference algorithms (graph-cuts, TRW-S, and belief propagation).
\cite{szeliski:pami08} considers a wider selection of problems---stereo matching, image reconstruction, photomontaging, and binary segmentation---but with 4-connectivity only, and applies a wider range of algorithms, adding ICM and $\alpha$-expansion to the above.
Recently, \cite{andres10dagm,kappes15ijcv} substantially widened the scope of such analysis, by considering also models with higher-order potentials, regular graphs with denser connectivity, models based on superpixels with smaller number of variables, and partitioning problems without unary terms. They compare the performance of many different types of algorithms on these models, including some specialised to particular problem classes.
These surveys help to understand the space of existing algorithms and provide a guide to which algorithms are suited for which models. Our work takes a natural step forward, with a technique to automatically select the best algorithm to run on an input problem instance.

\paragraph{Automatic algorithm selection.}
Automatic algorithm selection was pioneered by \cite{rice76ac}, which considered algorithms for quadrature and process scheduling.
More recently, machine learning techniques have been used to select algorithms for constraint-satisfaction~\cite{xu08jair}, and other combinatorial search problems~\cite{kotthoff11socs}.
However, none of these works consider selecting MAP inference algorithms. 

\section{Dataset of models}
\label{sec:dataset}

\paragraph{OpenGM2~\cite{kappes15ijcv}.}
The OpenGM2 dataset contains GMs drawn from 28 problem classes, including pairwise and higher-order models from computer vision and bioinformatics; it is the largest dataset of GMs currently available.
We briefly summarize here the main kinds of problems and refer to~\cite{kappes15ijcv} for details. 
\begin{itemize}[nosep]
\item
low-level vision problems such as stereo matching~\cite{szeliski:pami08}, inpainting~\cite{lellmann11jims,nowozin11iccv}, and montaging~\cite{szeliski:pami08}. These are all locally-connected graphs with variables corresponding to pixels, and with pairwise factors only; label counts vary widely between classes, from 2--256.
\item
small semantic segmentation problems with up to eight classes, with labels corresponding to surface types~\cite{gould09iccv} and geometric descriptions~\cite{hoiem11ijcv}. These are irregular, sparse graphs over superpixels; \cite{gould09iccv} uses pairwise factors only, while \cite{hoiem11ijcv} has general third-order terms.
\item
partitioning (unsupervised segmentation by clustering) based on patch similarity, operating on superpixels and with as many labels as variables, in both 2D~\cite{kim11nips,andres11iccv,brandes08kde} and 3D~\cite{andres12eccv_segmentation}. Potts or generalised Potts factors are used in all cases; \cite{kim11nips} has very large cliques with up to 300 variables, while the other classes are pairwise or third-order, just one class having dense connectivity.
\item
two problem classes from bioinformatics: protein side-chain prediction~\cite{jaimovich06jcb}, and protein folding~\cite{yanover08jcb}; both are defined over irregular graphs, with \cite{jaimovich06jcb} having only two labels but general third-order factors, while \cite{yanover08jcb} has up to 503 labels and dense pairwise connectivity.
\end{itemize}

\begin{figure}[t]
  \centering
  \renewcommand{\arraystretch}{1}
  \tabcolsep=0.05cm
  \begin{tabu}{cccc}
    \includegraphics[width=0.22\textwidth]{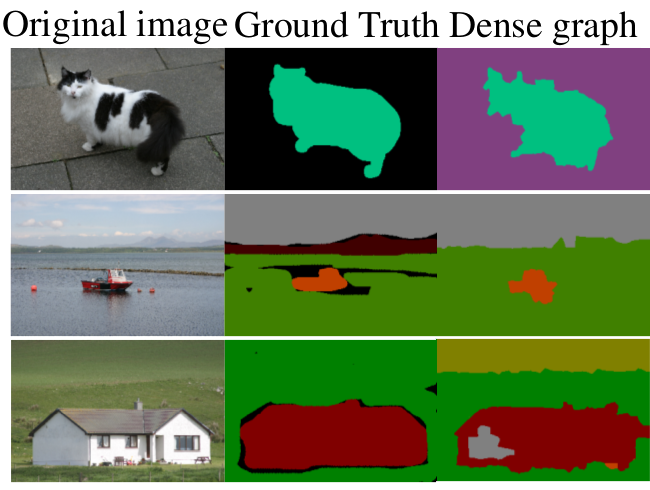} &
    \includegraphics[width=0.22\textwidth]{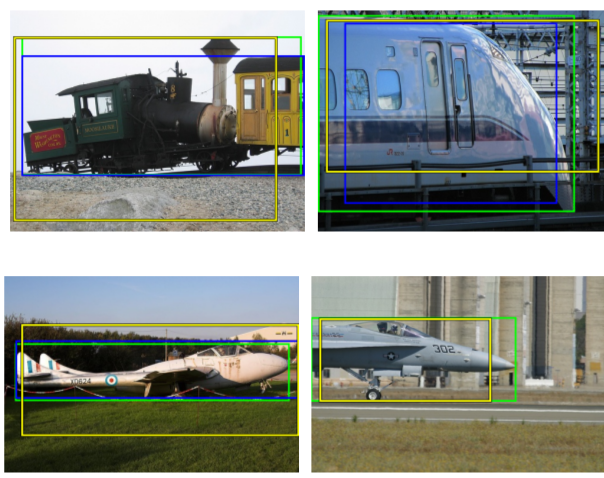} &
    \includegraphics[width=0.22\textwidth]{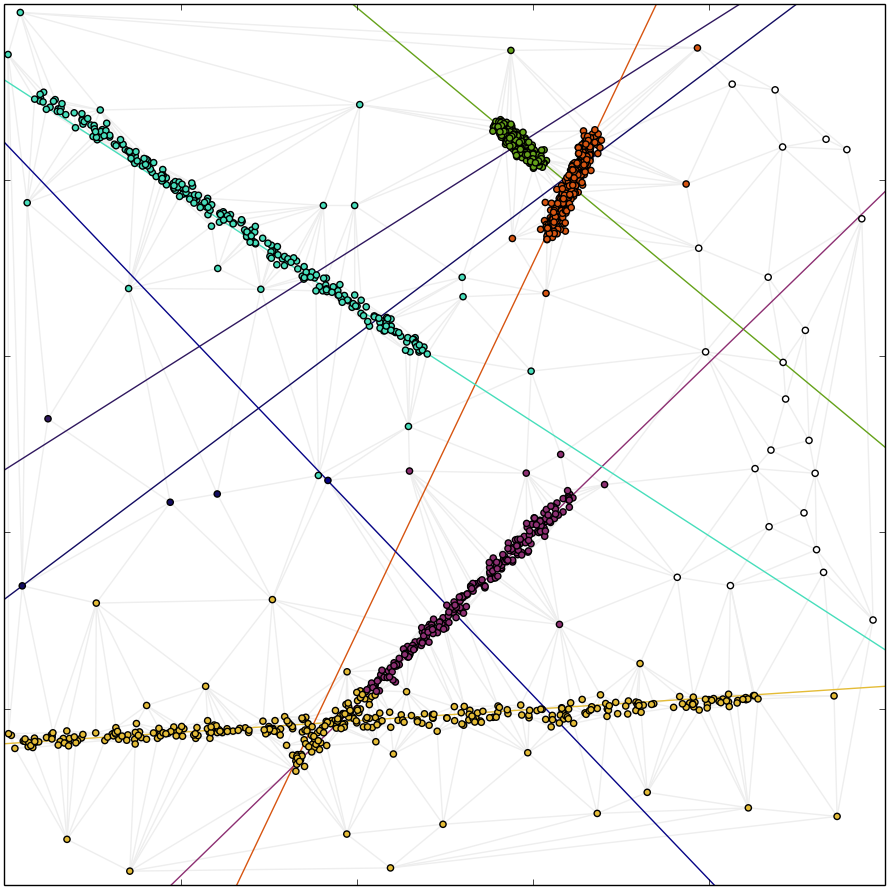} &
    \includegraphics[width=0.22\textwidth]{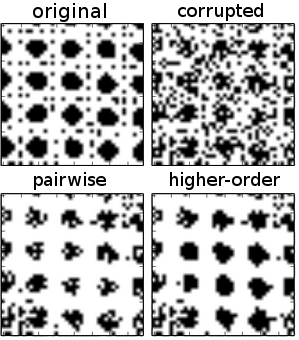} \\
    \rowfont{\fontsize{8pt}{10pt}\bfseries}
    msrc-segmentation &
    joint-localisation &
    line-fitting &
    texture-restoration \\
    \rowfont{\fontsize{8pt}{9pt}}
    $|V| = 275$ &
    $|V| = 81$ &
    $|V| = 661$ &
    $|V| = 1900$ \\
    \rowfont{\fontsize{8pt}{9pt}}
    $L = 21$ &
    $L = 100$ & 
    $L = 51$ &
    $L = 2$ \\
    \rowfont{\fontsize{8pt}{9pt}}
    $O = 2$ &
    $O = 2$ &
    $O = 2$ &
    $O = 49$
  \end{tabu}
  \renewcommand{\arraystretch}{\defaultarraystretch}
  \caption{Our four added GM classes. Variable count $|V|$ is a mean over all instances; $L$ is mean label count, and order $O$ is largest factor clique size}
  \afterfig
  \label{fig:added-crfs}
\end{figure}

\noindent Below we complement the OpenGM2 dataset with four additional, interesting problem classes which arise in modern computer vision applications (\fig{added-crfs}).

\paragraph{Semantic segmentation with context~\cite{guillaumin13cvpr}.}
Semantic segmentation on the MSRC-21 dataset~\cite{shotton09ijcv} with relative position factors.
Variables correspond to superpixels and labels to 21 object/background classes (\eg car, road, sky).
Unary factors are given by appearance classifiers on features of a superpixel, while pairwise factors encode relative location in the image, to favour labellings showing classes in the expected spatial relation to one another (\eg\ sky above road). The model is fully connected, \ie there is a pairwise factor between every two superpixels in the image.

\paragraph{Joint localisation~\cite{guillaumin13cvpr}.} 
Joint object localisation across images on the PASCAL VOC 2007 dataset~\cite{PASCAL07}. The set of images containing a certain object class form a problem instance.
Variables correspond to images and labels to object proposals~\cite{alexe10cvpr} in the images.
Unary factors are given by the objectness probability of a proposal~\cite{alexe10cvpr}, while pairwise factors measure the appearance similarity between two proposals in different images.
Inference on this model will select one proposal per image, so that they are likely to contain objects
and to be visually similar over the images.

\paragraph{Line fitting~\cite{isack12ijcv}.}
Fitting of multiple lines to a set of points in $\mathbb{R}^2$.
This an alternative to RANSAC~\cite{Fischler81} for fitting an unknown number of geometric models to a dataset.
Variables correspond to points and labels to candidate lines from a fixed pool (sampled from the point set in a preprocessing stage).
Unary factors favour labelling a point with a nearby line, while pairwise factors promote local smoothness of the labelling (\ie nearby points should take the same label). 

\paragraph{Texture restoration~\cite{rother09cvpr}.}
Binary texture restoration 
with pattern potentials. 
Given a binary image corrupted by noise, the task is to reconstruct the original noise-free image, while preserving the underlying texture regularity. 
Variables correspond to pixels and labels to `on' or `off'.
Unary factors penalise deviations from the input noisy image, while pairwise factors prefer pixels at certain offsets taking certain pairs of values (learned on a training image showing a noise-free texture).
Higher-order factors reward image patches for taking joint labellings which occur frequently in the training image (patterns).
The pairwise and higher-order factors capture low and high order texture properties, respectively.  

\paragraph{Data diversity.}
From each problem class we take all instances up to a maximum of 20. 
This results in a diverse dataset of 344 problem instances drawn from the 32 classes; 224 of these instances are pairwise and 120 higher-order.
21 of the problem classes have small label-spaces ($<20$ labels), while the remainder vary greatly up to a maximum of 17074.
Variable counts similarly cover a wide range, from 19 to 2356620, with a median of 10148.
Amongst the higher-order problems, 58\% of instances have arbitrary dense factor tables, while the remainder have Potts potentials~\cite{boykov01pami} or generalised versions thereof~\cite{kohli07cvpr,KohliCVPR08}.
The problem classes also differ greatly in the degrees of homogeneity of their instances. For example, instances in the \textit{line-fitting} class vary by an order of magnitude in variable and label counts, whereas all instances in the \textit{inclusion} class have identical characteristics but for the factor energies themselves.

\section{Inference algorithms and performance}
\label{sec:algorithms}

\paragraph{Inference algorithms.}
A vast number of MAP inference algorithms have been proposed in the literature, with differing approaches, degrees of generality, and performance characteristics.
We selected 15 to use in our experiments (\tab{algorithm-list}),
including representative algorithms from most prominent approaches, \eg move-making, message-passing, dual-decomposition, combinatorial, etc.
This covers many of the most commonly used algorithms in computer vision, such as TRW-S~\cite{kolmogorov06pami}, QPBO~\cite{rother:cvpr07}, and $\alpha$-expansion~\cite{boykov01pami}.
Note however that we do not aim to form an exhaustive pool of all good algorithms; our automated selection method is agnostic to the pool of algorithms it is trained to select from, and explicitly avoids making prior assumptions on their applicability.

\begin{table}[t]
  \small
  \centering

  \caption{\small Algorithms used in this study, including the GM orders they are applicable to (pw = pairwise), number of parameter settings included if more than one (\#p), full name or description, and reference to the original work}
  \label{tab:algorithm-list}

  \renewcommand{\arraystretch}{1.1}
  \tabcolsep=0.05cm
  \begin{tabu}{ccc@{\raggedright\hskip 0.1cm}X[-1,m]c}
  	\rowfont\bfseries
    alias & order & \#p & name / description & ref. \\
    \tabucline \\
    A* & {\em all} & & implicitly convert to shortest-path problem and apply A* & \cite{Bergtholdt08} \\
    AD\textsuperscript{3} & {\em all} & & alternating directions dual decomposition with branch and bound & \cite{martins15jmlr} \\
    $\alpha$-exp & \em{pw} & & alpha-expansion & \cite{boykov01pami} \\
    BPS & {\em all} & 4 & sequential loopy belief propagation, implementation of \cite{kappes15ijcv} & \cite{Bishop-book} \\
    DDS & {\em all} & 2 & dual decomposition with subgradient descent & \cite{kappes12cvpr}\\
    FPD & {\em pw} & 3 & fast primal/dual (FastPD) & \cite{komodakis08cviu} \\
    ICM & {\em all} & & iterated conditional modes & \cite{besag86jrss} \\
    ILP & {\em all} & & solve as integer programming problem with Gurobi & \cite{kappes15ijcv} \\
    KL & {\em pw} & & Kernighan-Lin method for 2\textsuperscript{nd} order partitioning problems & \cite{kernighan70bstj} \\
    LBP & {\em all} & 4 & parallel loopy belief propagation, implementation of \cite{kappes15ijcv} & \cite{Bishop-book} \\
    LP & {\em all} & & solve linear programming relaxation with Gurobi & \cite{kappes15ijcv} \\
    MPLP & {\em all} & & max-product linear programming with cutting plane relaxation tightening & \cite{sontag08uai,sontag12uai} \\
    QPBO & {\em pw} & & quadratic pseudo-boolean optimisation & \cite{rother:cvpr07} \\
    TRW-S & {\em pw} & 3 & sequential tree-reweighted message-passing & \cite{kolmogorov06pami} \\
    UM & {\em all} & & take lowest-energy label according to unary factors only & -
  \end{tabu}
  \afterfig
\end{table}

We also include a simple method, dubbed {\em unary-modes (UM)}, which labels each variable by minimizing its unary factors only; this should perform poorly on genuinely hard structured prediction problems, where the non-unary factors have a decisive impact on the MAP labelling.

\paragraph{Protocol for inference.}
We used the original authors' implementation of each algorithm where available, and the implementations in \cite{kappes15ijcv} otherwise.
Every algorithm was run on every problem instance in our dataset, with limits of 60 minutes CPU time and 4GB RAM imposed for inference on one instance.
For each successful run, we recorded the MAP labelling and time taken.

Many of the algorithms have free parameters that must be defined by the user.
While it was not practical to evaluate every possible combination of parameters, for several of the algorithms we included multiple parameterisations where this affects their results significantly.
For example, we ran four versions of loopy belief propagation, with damping set to 0.0 and 0.75, and maximum iteration counts of 50 and 250.
In such cases, the different parameterisations are combined to create a meta-algorithm, which
simulates the user running every parameterisation, then taking the results from that yielding lowest energy on the problem instance.

Several incompatible combinations of algorithms and GMs were included. When possible, we still ran the algorithm to obtain an approximate solution:
\begin{itemize}[nosep]
\item
higher-order factors are omitted when passing GMs to pairwise algorithms. However, when evaluating the algorithm's performance, the energy of the output labelling is still computed on the full model including all factors.
\item
non-metric pairwise factors passed to $\alpha$-expansion are handled as if they were metric, sacrificing the usual correctness and optimality guarantees~\cite{boykov01pami}.
\end{itemize}
When it was not possible to run the algorithm, we counted this as a failure:
\begin{itemize}[nosep]
\item
QPBO aborts when presented with a GM having non-binary variables.
\item
FastPD aborts when presented with a GM whose pairwise factors are not all proportional to some uniform distance function on labels.
\item
Kernighan-Lin aborts when presented with a GM having factors that are not pairwise Potts
\end{itemize}

\begin{table}[t]
  \small
  \centering

  \caption{Aggregate performance of each inference algorithm on our dataset; mean time is over instances for which the algorithm successfully returns a result}
  \label{tab:algo-true-performance}

  \renewcommand{\arraystretch}{1.1}
  \tabcolsep=0.1cm
  \begin{tabu}{cccc@{\hskip 0.4cm}c}
    \rowfont{\bfseries}
     & \multicolumn{3}{c}{\% instances for which...} & \multirow{2}{*}{mean time /s} \\[-2pt]
    \rowfont{\itshape}
     & completes & best-\&-fastest & good-\&-fastest \\
    \tabucline \\
A*                    &  4 &  0 &  0 & 0.1 \\
AD\textsuperscript{3} & 52 &  7 &  1 & 390.2 \\
$\alpha$-exp          & 98 &  5 &  7 & 23.4 \\
BPS                   & 72 &  4 &  2 & 158.3 \\
DDS                   & 80 &  0 &  0 & 296.6 \\
FPD                   & 31 &  9 & 22 & 7.2 \\
ILP                   & 48 &  1 &  0 & 96.3 \\
LP                    & 52 &  2 &  1 & 76.8 \\
ICM                   & 100 & 30 & 31 & 60.7 \\
KL                    & 12 & 10 & 10 & 142.2 \\
LBP                   & 73 &  6 &  4 & 193.5 \\
MPLP                  & 56 &  1 &  1 & 1116.3 \\
QPBO                  & 12 &  0 &  2 & 0.1 \\
TRW-S                 & 94 & 19 & 10 & 236.4 \\
UM                    & 100 &  0 &  3 & 0.1
  \end{tabu}
  \renewcommand{\arraystretch}{\defaultarraystretch}
  \afterfig
\end{table}

\paragraph{Performance measures.}
We measured three aspects of the performance of each algorithm:
\begin{itemize}[nosep]
\item {\em completes:}
whether the algorithm runs to completion, \ie returns a solution within 60 minutes, regardless of the energy of that solution.
\item {\em best-and-fastest:}
whether the algorithm reaches the lowest energy among all algorithms, faster than any other one that does so.
This is relevant for a user requiring the solution with lowest possible energy, even at high computational cost.
\item {\em good-and-fastest:}
whether the algorithm is the fastest to reach a solution with 98\% of variables matching the lowest energy labelling.
This is highly relevant in practice, as minor deviations from that labelling may not matter to the user, while achieving it would require a significantly slower algorithm.
%
\end{itemize}

\noindent Table \ref{tab:algo-true-performance} shows the performance of the algorithms with respect to these measures.


\paragraph{Algorithm diversity.}
We see that the distributions of both best-and-fastest and good-and-fastest algorithms over instances have high entropy---many different algorithms are best-and-fastest or good-and-fastest for a significant fraction of GMs.
11 of the 15 algorithms are able to return a solution for at least one instance on more than half of the problem classes; the other four are particularly restricted, such as QPBO (which only operates on binary problems).
All the algorithms other than A* and DDS are the best-and-fastest for at least one problem instance.
TRW-S and FastPD perform particularly well on pairwise problems, with TRW-S generally reaching slightly lower energies, but FastPD being much quicker. Kernighan-Lin outperforms all algorithms on pairwise partitioning problems.
AD\textsuperscript{3} gives low energies for high-order problems, but often takes longer than other algorithms.
Only ICM and unary-modes are able to return a solution for all problem instances. Although they are fast and widely-applicable, these na\"{i}ve methods are unable to return the best solution in the majority of cases. 
%
All these observations show how our goal of learning to select the best inferencer is much harder than simply picking any algorithm that runs to completion.

\section{Learning to select an algorithm}
\label{sec:learnt-selection}

We now consider how to automatically select the best MAP inference algorithm for an input problem instance. This is the main contribution of this paper.
We define two tasks:
(1) predicting the best-and-fastest algorithm;
and
(2) predicting the good-and-fastest algorithm.
To address these tasks, we design selection models that take a GM as input, and select an algorithm as output (\sect{models}).
The selection models operate on features extracted from the GMs themselves (\sect{features}).
This is different from the typical approach in computer vision of extracting features from images and using these to build a GM.

\subsection{GM features}\label{sec:features}

\begin{figure}[t]
  \centering
  \begin{subfigure}{0.28\textwidth}
    \centering
    \includegraphics[scale=1.25]{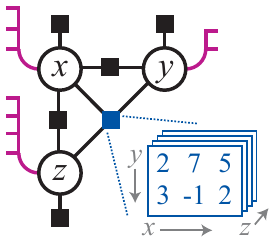}
    \caption{}
    \label{fig:example-features-model}
  \end{subfigure}~
  \begin{subfigure}{0.16\textwidth}
    \centering
    $|V| = 3$ \\[4pt]
    $|F| = 6$ \\[4pt]
    $L_\textit{min} = 2$ \\
    $L_\textit{max} = 4$\\
    $L_\textit{mean} = 3$
    \caption{size}
    \label{fig:example-features-counts}
  \end{subfigure}~
  \begin{subfigure}{0.26\textwidth}
    \centering
    \includegraphics[width=0.5\textwidth]{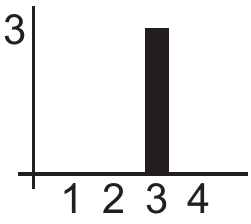}
    \\
    \includegraphics[width=0.5\textwidth]{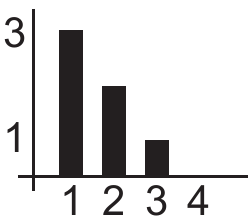}
    \caption{variable (top) and factor (bottom) orders}
    \label{fig:example-features-orders}
  \end{subfigure}~
  \begin{subfigure}{0.25\textwidth}
    \centering
    \tabcolsep=0.1cm
    \tabulinesep=0.1cm
    \begin{tabu}{c|c}
      \rowfont\bfseries
      \small order & \small density \\[-2pt]
      \tabucline \\
      2 & $\dfrac{2}{\binom{3}{2}} = \frac{2}{3}$ \\
      3 & $\dfrac{1}{\binom{3}{3}} = 1$ \\
      4 & 0 \\
    \end{tabu}
    \caption{factor densities}
    \label{fig:example-features-factor-densities}
  \end{subfigure}
  \caption{\small Example GM structure (a) and associated features (b-e). Circles correspond to variables, and squares to factors; the label space of each variable is shown as purple dashes. Part of the value table for the third-order factor (blue) is also shown}
  \label{fig:example-features}
  \afterfig
\end{figure}

We extract the following three groups of features from each problem instance (\fig{example-features}).

\paragraph{Instance size.}
The number of variables, $|V|$, and of factors, $|F|$, are used to indicate the overall size of the problem instance, hence whether slower algorithms are likely to be applicable.
We also include the minimum, maximum and mean label count over all variables. See \fig{example-features-counts}.
    
\paragraph{Structural features.}
We extract more sophisticated features based on the model structure, \ie which do not depend on the factor values themselves.
Firstly, we take a histogram and statistics (minimum, maximum, mean) of both:
\begin{itemize}[nosep]
  \item variable orders (\ie for each variable, number of connected factors, \fig{example-features-orders})
  \item factor orders (\ie for each factor, number of variables in its clique, \fig{example-features-orders})
\end{itemize}
Secondly, we measure factor densities---for each factor order $M \geq 2$, the number of factors of order $M$ divided by the binomial coefficient $\binom{|V|}{M} = \frac{|V|!}{M!(|V|-M)!}$.
Intuitively, this is the number of possible $M$-cliques that actually have an associated factor.
In \fig{example-features}, this is 1 for third order, as there is only one possible triplet, but $2/3$ for second order, as only two of the possible three pairs of variables have a pairwise factor: $(x,y)$ and $(x,z)$ but not $(z,y)$
  
\paragraph{Energy features.}
Our final group of features depend on the values of the factors themselves, \eg the blue values in \fig{example-features-model}.
To determine the influence of different orders of factors, we compute means and deviations over values they take, defined as follows:
\begin{itemize}[nosep]
\item
  for each factor $f \in F$, let $\mu_f$ be the mean and $\sigma_f$ the standard deviation of all unique values taken by $f$
\item
  then, for each factor order $M \geq 2$, compute for factors $F_M$ of that order, the ratio of each of the following to the same quantity for $M = 1$: \\[3pt]
  (i)~~$\sum_{f \in F_M}\mu_f$ \hskip 16pt
  (ii)~~$\sum_{f \in F_M}\mu_f / |F_M|$ \hskip 16pt
  (iii)~~$\sum_{f \in F_M}\sigma_f / |F_M|$.
\end{itemize}
Intuitively, these capture how much influence each order of factor has on the final energy, \ie how much changing the labelling will change the values of factors of each order. On pairwise GMs, a large influence of the pairwise (as opposed to unary) factors makes inference harder; on higher-order GMs, pairwise algorithms should perform relatively well only when the influence of higher-order factors is small.
We also count the fraction of pairwise factors $f$ having each of the following characteristics for all labels $a$, $b$, $c$:\\
\begin{tabu}{@{\hskip 16pt}cl@{\hskip 16pt}cl}
  (i) & $f(a, b) = 0$ iff $a = b$ &
  (ii) & $f(a, b) \geq 0$ \\
  (iii) & $f(a, b) = f(b, a)$ &
  (iv) & $f(a, b) + f(b, c) \geq f(a, c)$.
  \end{tabu}
\\
Together, these are the conditions for a factor to be metric; without (iv), to be semi-metric---respectively requirements for the $\alpha$-expansion and $\alpha\beta$-swap algorithms to fulfill their correctness guarantees~\cite{boykov01pami}.
Finally, we measure the fraction of pairwise factors which are submodular; in general pairwise submodular problems are easier to solve by LP-based methods, as their LP relaxation is tight.

\subsection{Algorithm selection models and their training}
\label{sec:models}

\paragraph{Selection models.}
We propose two algorithm selection models.
Each is a 1-of-N classifier implemented as a random forest~\cite{criminisi2011}, taking the features described in \sect{features} as input.
Model {\em BF} is trained to predict the best-and-fastest algorithm; model {\em GF} is trained to predict the good-and-fastest algorithm.
The random forests are trained recursively by selecting the best split from a randomly-generated pool at each step, using information gain (\ie entropy decrease) as the criterion, and with outputs modelled by categorical distributions \cite{criminisi2011}.
\paragraph{Data.}
We train the selection models on a subset of our dataset (\sect{dataset}). A training sample consists of features extracted from a problem instance and a target output label denoting which algorithm works best on it. It is important to note that these training labels are automatically generated by running all algorithms on the training instances, as in \sect{algorithms}. No human annotation is required.
At test time, we run the selection models on a separate subset of the dataset. The evaluation compares the algorithm selected by our model to the one known to perform best (\sect{results}). Again, this test label is produced automatically.

\section{Experiments}
\label{sec:results}

\paragraph{Tasks and baselines.}
We report results on the two tasks defined in \sect{learnt-selection}: (i) predicting the good-and-fastest inference algorithm for an input GM; and (ii) predicting the best-and-fastest algorithm.
Task (i) is addressed by the selection model GF, and task (ii) by model BF (\sect{models}).
For both tasks, we also analyse the performance of two baseline methods that select an algorithm without looking at features of the input instance.
The first baseline \textit{NB} always selects the algorithm that is most often best over the full training set.
This mimics the behaviour of a na\"{i}ve user who simply chooses one commonly good algorithm to use.
For the second, stronger baseline \textit{SB}, we assign each of the problem classes to one of three superclasses:
\begin{enumerate}[nosep]
\item pairwise---many algorithms are designed for pairwise problems only;
\item higher-order---there exist algorithms designed explicitly to handle higher-order factors, but which may be slow for pairwise instances;
\item partitioning---these are a special class which is hard for general algorithms (due to having a large label space, and being invariant to label permutations) but certain methods can exploit this structure to solve them efficiently; most partitioning problems in our dataset are pairwise, but some are third-order.
\end{enumerate}
Then, at test time, each problem instance is assigned the algorithm that is most often best for training problems of its superclass.
This strong baseline mimics the behaviour of a user with good working knowledge of inference---enough to recognise how her problem fits in these superclasses, and to know which algorithm will be best for each.
%

\paragraph{Experimental setup.}
We select half the problem instances at random to train on, and the remainder are used for testing.
As discussed in \sect{algorithms}, the ground-truth labels marking which inference algorithms perform best on a problem instance are obtained automatically by running all algorithms on all instances. No human annotation is necessary for training or testing.

When training and evaluating selection models, the underlying problem classes are always treated as unknown---they are not provided as input data.
We want the selection models to freely learn the optimal association between GM features and good algorithms to run. The GM features we propose are designed to enable the selection models to reason upon various properties of GMs, which can be used to characterize problem classes (\eg connectivity structure and distributions of energy values in the factors).
So, we might expect the selection models to learn at least some of the problem class structures, given that this often correlates with the best algorithms (\sect{algorithms}).

\begin{table}[t]
  \centering
  \small

  \caption{\small Performance of our model GF and baselines NB and SB for selecting the good-and-fastest algorithm (first three columns), and performance of our model FG and baselines for selecting the best-and-fastest algorithm (last three columns).
  }
  \label{tab:aggregates-eqs}

    \renewcommand{\arraystretch}{1.15}
    \tabcolsep=0.09cm
    \begin{tabu}{r@{\hskip 0.3cm}ccc@{\hskip 0.4cm}ccc}
      \rowfont\itshape
       & \multicolumn{3}{c}{good-and-fastest} \hskip 0.4cm & \multicolumn{3}{c}{best-and-fastest} \hskip 0.3cm \\[-2pt]
      \rowfont\bfseries
      algorithm selected by... & GF & NB & SB & BF & NB & SB \\
      \tabucline\\ \\[-12pt]

      \% instances correctly classified &
        69 & 31 & 28 & 62 & 30 & 36 \\ 
      mean \% matching variables &
        96.4 & 75.3 & 87.5 & 97.1 & 75.4 & 95.6 \\[2pt] 

    \end{tabu}
    \afterfig
\end{table}

\paragraph{Evaluation measures.}
Our algorithm selection models are evaluated on the test set with the following measures (\tab{aggregates-eqs}):
\begin{itemize}[nosep]
\item percentage of instances with the correct algorithm (best-and-fastest or good-and-fastest) selected. This is the measure for which we trained our selection models.
%
%
\item mean (over instances) of fraction of variables matching the labelling returned by the best-and-fastest algorithm.
This is particularly relevant in practice, as users typically care about the quality of the labelling output, by an algorithm, especially in terms of how close it is to the lowest-energy labelling that could have been returned.
\end{itemize}

\section{Analysis}

\paragraph{Predicting the good-and-fastest algorithm.}
Model GF correctly chooses the good-and-fastest algorithm for 69\% of instances, with 96.4\% of variables taking the same label as in the true best labelling on average. 
This compares favourably to the na\"{i}ve baseline NB, which correctly selects only on 31\% of the instances and returns labellings that are considerably worse (75.3\% correctly-labelled variables on average).
Indeed, our model also substantially outperforms the strong baseline, which only achieves an average of 87.5\% of correctly-labelled variables.

These results show that our selection model successfully generalises to new problem instances not seen during training. It is able to select an algorithm much better than even the strong baseline of a user who knows which algorithm performs best for similar problems in the training set.

\paragraph{Predicting the best-and-fastest algorithm.}
Model BF correctly selects the best-and-fastest algorithm on 62\% of instances, exceeding the na\"{i}ve baseline (32\%).
This results in 97.1\% of variables taking the same label as in the true lowest-energy solution, greatly exceeding the na\"{i}ve baseline of 75.4\%.
Our model performs well against even the strong baseline, which only classifies 36\% of instances correctly and has a slightly lower fraction of correct variables at 95.6\%.

\begin{table}[t]
  \small
  \centering

  \caption{\small Mean times and speed-ups from using our method, versus exhaustively applying all algorithms. \textit{matching var's} is fraction of variables whose labels match true best result; \textit{speed-up} is ratio of time to that for exhaustive testing}
  \label{tab:efficiency}

  \tabcolsep=0.15cm
  \begin{tabu}{rccc}
    \rowfont\bfseries
    mean... & time /s & speed-up & matching var's \\
    \tabucline \\
    exhaustive & 13046.8 & $1.0\times$ & 100\% \\[-1pt]
    good-and-fastest & 221.3 & $88.1\times$ & 96.4\% \\[-1pt]
    best-and-fastest & 312.5 & $46.8\times$ & 97.1\%
  \end{tabu}
  \afterfig
\end{table}

\paragraph{Efficiency.}
As noted in \sect{introduction}, a simple alternative to our selection method is to run every algorithm on the test problem instance, and select the lowest-energy solution.
However, this is computationally very expensive.
To evaluate the speed-up made by our method, for each problem instance we also measured (i) the total time to run every inference algorithm; (ii) the time to predict the best-and-fastest algorithm with model BF then run it; and, (iii) the time to predict the good-and-fastest algorithm with model GF then run it.
As we see in \tab{efficiency}, our method results in an average speed-up of $46.8\times$ using model BF, and $88.1\times$ using model GF, with 97.1\% and 96.4\% of variables correctly labelled respectively.
Thus, automated selection achieves labellings very similar to running every algorithm, but at a small fraction the computational expense.
Model GF yields a significantly faster-running algorithm on average than model BF, with only a small drop ($< 1\%$) in variables correctly labelled.

\begin{table}[t]
  \fontsize{8pt}{9pt}
  \centering

  \caption{\small Confusion matrix showing true (rows) and predicted (columns) good-and-fastest algorithms for pairwise problems. The table only includes those algorithms that are the true good-and-fastest for at least one problem instance.}
  \label{tab:pairwise-algo-confusion}

  \renewcommand{\arraystretch}{1.12}
  \tabcolsep=0.1cm
  \begin{tabu}{>{\bf}r|cccccccccc}
    \rowfont\bfseries
 & AD\textsuperscript{3} & $\alpha$-exp & BPS & FPD & ICM & KL & LBP & QPBO & TRW-S & UM \\
\tabucline \\
AD\textsuperscript{3} & \cellcolor{gray!25} 0 & 0 & 0 & 0 & 2 & 0 & 0 & 0 & 0 & 0 \\
$\alpha$-exp & 0 & \cellcolor{gray!25} 5 & 1 & 1 & 0 & 0 & 0 & 0 & 2 & 0 \\
BPS & 0 & 0 & \cellcolor{gray!25} 1 & 0 & 2 & 0 & 1 & 0 & 1 & 0 \\
FPD & 0 & 0 & 1 & \cellcolor{gray!25} 19 & 0 & 0 & 0 & 0 & 0 & 0 \\
ICM & 0 & 0 & 0 & 0 & \cellcolor{gray!25} 16 & 1 & 0 & 0 & 3 & 0 \\
KL & 0 & 0 & 0 & 0 & 0 & \cellcolor{gray!25} 24 & 0 & 0 & 0 & 0 \\
LBP & 0 & 0 & 0 & 0 & 3 & 0 & \cellcolor{gray!25} 4 & 0 & 0 & 1 \\
QPBO & 0 & 0 & 3 & 0 & 0 & 0 & 0 & \cellcolor{gray!25} 3 & 0 & 0 \\
TRW-S & 0 & 1 & 0 & 6 & 1 & 0 & 1 & 0 & \cellcolor{gray!25} 6 & 0 \\
UM & 0 & 1 & 0 & 1 & 0 & 0 & 0 & 1 & 0 & \cellcolor{gray!25} 0
  \end{tabu}
  \renewcommand{\arraystretch}{\defaultarraystretch}
  \afterfig
\end{table}

\paragraph{Algorithms selected by the strong baseline.}
As described in \sect{results}, our strong baseline chooses the algorithm that is most often best-and-fastest or good-and-fastest over the training set, for problems in the same superclass as the test instance.
For predicting the best-and-fastest algorithm, Kernighan-Lin is selected for partitioning problems, TRW-S for other pairwise instances, and AD\textsuperscript{3} for other (higher order) instances.
However, for the good-and-fastest algorithm, FastPD is selected instead of TRW-S for pairwise instances, and ICM for higher order instances, indicating that these often label 98\% or more of variables correctly, while being faster to run.

\paragraph{Algorithms selected by our method.}
%
%
At a coarse level, for the task of selecting the best-and-fastest algorithm, we find that our model BF most often chooses pairwise-specific algorithms for pairwise problems, and AD\textsuperscript{3} for higher-order problems.
This agrees with intuition---pairwise algorithms are specifically designed to be faster for pairwise instances, while AD\textsuperscript{3} is a good general-purpose algorithm for higher-order instances.
Interestingly, for the good-and-fastest task, model GF correctly learns to choose ICM or a good pairwise method for higher-order problems in place of AD\textsuperscript{3}---for many instances, these provide solutions close in labelling to the lowest-energy, and do so much faster than AD\textsuperscript{3}.

%
To explore whether our method can also draw more subtle distinctions,
we now examine the distribution of algorithms it selects for pairwise problems.
10 of the algorithms we consider are useful for these, in the sense of being good-and-fastest for at least one instance.
Table \ref{tab:pairwise-algo-confusion} shows the confusion matrix for true and predicted good-and-fastest algorithms amongst these 10.
Certain groups of problems can be distinguished based on structural properties, such as partitioning problems to be solved with KL, or very large instances that only run to completion with ICM. Our model correctly makes these distinctions.
Other distinctions are even more subtle---such as whether to use $\alpha$-expansion, TRW-S, or FastPD for a pairwise problem of moderate size.
Our model is able to select between these three algorithms, making the correct choice for 75\% of instances.

\begin{table}[t]
  \centering
  \small

  \caption{\small Performance of algorithm selection methods selecting good-and-fastest and best-and-fastest algorithms in the LOCO regime; see \tab{aggregates-eqs} for details.
  }
  \label{tab:aggregates-loco}

    \renewcommand{\arraystretch}{1.15}
    \tabcolsep=0.09cm
    \begin{tabu}{r@{\hskip 0.3cm}ccc@{\hskip 0.4cm}ccc}
      \rowfont\itshape
       & \multicolumn{3}{c}{good-and-fastest} \hskip 0.4cm & \multicolumn{3}{c}{best-and-fastest} \hskip 0.3cm \\[-2pt]
      \rowfont\bfseries
      algorithm selected by... & GF & NB & SB & BF & NB & SB \\
      \tabucline\\ \\[-12pt]

      \% instances correctly classified &
        40 & 26 & 11 & 28 & 25 & 23 \\ 
      mean \% matching variables &
        89.8 & 73.3 & 71.7 & 85.5 & 73.3 & 86.2 
    \end{tabu}
    \afterfig
\end{table}

\paragraph{LOCO regime.}
We also tested our models and baselines in an even harder `leave one class out' (LOCO) regime, where for each problem class $C$ in turn, we train on all instances from classes other than $C$, and test on those in $C$; the final performance is given by a weighted mean over classes.
This tests generalisation to classes absent from the training set, which is relevant when the user does not wish to train our model on her classes.
The results are presented in \tab{aggregates-loco}.

For selecting the good-and-fastest algorithm, model GF still performs well in LOCO regime, selecting algorithms labelling 89.8\% of variables correctly, and exceeding both na\"{i}ve and strong baselines by over 15\%.
Moreover, we correctly choose the good-and-fastest algorithm 14\% more often than the baselines.

For the best-and-fastest task, model BF results in 85.5\% of variables being correctly labelled, significantly exceeding the na\"{i}ve baseline at 73.3\% and comparable with the strong baseline at 86.2\%.
These results demonstrate that our selection models are strong enough to generalise \textit{across} the hidden problem classes, going beyond discovering and recalling distinguishing features of these. 

\section{Conclusions}\label{sec:conclusion}

We have presented a method to automatically choose the best inference algorithm to apply to an input problem instance.
It selects an inference algorithm that labels 96\% of variables the same as the best available algorithm for that instance.
Our method is over $88\times$ faster than exhaustively trying all algorithms. The experiments show that our automated selection methods successfully generalise across problem instances and importantly, even across problem classes.

\bibliographystyle{splncs}
\bibliography{../../bibtex/shortstrings,../../bibtex/calvin,../../bibtex/vggroup}

\end{document}